\pdfoutput=1

\documentclass[11pt,table]{article}

\usepackage[table,xcdraw]{xcolor}
\usepackage[]{EMNLP2022}

\usepackage{times}
\usepackage{latexsym}

\usepackage[T1]{fontenc}

\usepackage[utf8]{inputenc}

\usepackage{microtype}

\usepackage{inconsolata}

\usepackage{cjhebrew}
\usepackage{linguex}
\usepackage{leipzig}
\usepackage{ucs}
\usepackage{graphicx}
\usepackage[caption=false]{subfig}
\usepackage{comment}

\newcommand{\corpname}{\textsc{IAHLTwiki}}

\title{A Second Wave of UD Hebrew Treebanking and Cross-Domain Parsing}

\author{
  Amir Zeldes \\
  Georgetown University \\
  \texttt{amir.zeldes@georgetown.edu} \\\And
  Nick Howell \\
  IAHLT \\
  \texttt{nlhowell@gmail.com} \\\AND
  Noam Ordan \\
  IAHLT \\
  \texttt{noam.ordan@gmail.com} \\\And
  Yifat Ben Moshe \\
  IAHLT \\
  \texttt{yifat1811@gmail.com} \\
  }

\begin{document}
\maketitle
\begin{abstract}
Foundational Hebrew NLP tasks such as segmentation, tagging and parsing, have relied to date on various versions of the Hebrew Treebank (HTB, \citealt{SimaanItaiWinterEtAl2001}). However, the data in HTB, a single-source newswire corpus, is now over 30 years old, and does not cover many aspects of contemporary Hebrew on the web. This paper presents a new, freely available UD treebank of Hebrew stratified from a range of topics selected from Hebrew Wikipedia. In addition to introducing the corpus and evaluating the quality of its annotations, we deploy automatic validation tools based on grew \cite{Guillaume2021}, and conduct the first cross-domain parsing experiments in Hebrew. We obtain new state-of-the-art (SOTA) results on UD NLP tasks, using a combination of the latest language modelling and some incremental improvements to existing transformer based approaches. We also release a new version of the UD HTB matching annotation scheme updates from our new corpus.
\end{abstract}

\section{Introduction}

Treebanks (TBs) form a fundamental resource for NLP and computational linguistics research: they provide high quality annotated data for tokenization, sentence splitting, POS tagging and syntactic/semantic relation extraction. For Morphologically Rich Languages (MRLs, \citealp{SeddahKueblerTsarfaty2014}), high quality morphosyntactically annotated data is particularly crucial, since basic search on a word-matching level cannot work before morphological segmentation has occurred. In languages such as Hebrew, where vowels are not well represented in the script, the need for reliable morphosyntactic disambiguation is particularly strong, since string-level ambiguity is very frequent and high. This is demonstrated by the following frequently cited example from \citet{AdlerElhadad2006}, which has a large number of possible analyses for just a four character sequence (note Hebrew is right-to-left).

\exg. \begin{cjhebrew}Ml.sb\end{cjhebrew}  \\
  {\small $\langle$b.cl.m$\rangle$ ~~~be.cil.am - in.shadow.their }\\
  {\small $\langle$b.clm$\rangle$ \quad (be./b.a.)celem - in.(a/the).image}\\
  {\small $\langle$b.clm$\rangle$ \quad (be./b.a.)calam - in.(a/the).photographer}\\
  {\small $\langle$bcl.m$\rangle$ \quad bcal.am - onion.their}\\
  {\small $\langle$bclm$\rangle$ \quad  ~becelem - Betzelem (organization)}\label{bclm}

\vspace{-3pt}

When such sequences contain multiple sub-tokens (e.g.~`in' and `the' within `in.the.image'), we follow  Universal Dependencies\footnote{\url{https://universaldependencies.org/}} terminology in referring to the larger unit as a Multi-Word Token (MWT), and the sub-parts, each of which carries a separate part-of-speech, as tokens.

Although a treebank for Hebrew was already created by the MILA Institute (\citealt{SimaanItaiWinterEtAl2001}, hence HTB), subsequently converted to dependencies \cite{goldberg-elhadad-2009-hebrew} and finally to the more recent Universal Dependencies framework (see \citealt{sade-etal-2018-hebrew}, hence UD-HTB), two major concerns motivate the current work in creating a new UD treebank for Hebrew. The first is the age of the data: HTB texts were taken from 1990--91 issues of a single newspaper, Ha'aretz, as illustrated in \ref{ex:haaretz}, describing the introduction of computers to an office.
\vspace{-3pt}

\ex. \textit{ba-axrona huxnesu ma`arexet maxshev ve-toxnot le-kol ha`anafim} \\
    `Recently a computer system and software were introduced into all branches'\label{ex:haaretz}

\vspace{-3pt}
This is one of only a dozen examples of the word `computer' in the data, which predates mainstream Web 1.0 times, mentions no cellular phones, the Internet, or a variety of countries established after 1990, including the EU. Previous work has shown that NLP systems retain strong lexical biases mirroring both period and author demographics \cite{shah-etal-2020-predictive}, which in the case of HTB reflect primarily Israeli Ha'aretz journalists from 1990. 

A second concern beyond the period and author demographics is genre. HTB is a news corpus, and as such reflects formal journalistic writing, which focuses on describing prominent political events, sports news, and reported speech, usually in the past tense, but under-represents expository text, academic language, and colloquial spellings (which are much more variable in Hebrew than in English). The importance of genre diversity in training data has often been noted \cite{ZeldesSimonson2016,MuellerEbersteinEtAl2021}, but without a second genre to test on, we simply do not know the extent to which fundamental Hebrew NLP performance degrades outside of HTB's language.

In this paper, we attempt to broaden the range of data available for Hebrew by creating and evaluating a new, freely available (CC-BY-SA license), gold standard corpus using contemporary Wikipedia data. Although Wikipedia's language is also relatively formal, it differs substantially from newspaper reporting, and is more contemporary than HTB, covering a broad range of topics, while being available under an open license.\footnote{We also note the current development of UD resources for Biblical Hebrew (\url{https://github.com/UniversalDependencies/UD_Ancient_Hebrew-PTNK}), however we concur with UD in classifying ancient varieties as separate languages.} Our main contributions are:
\vspace{-5pt}

\begin{enumerate}
\itemsep0.07em 
    \item A new TB of Hebrew Wikipedia data from several domains annotated for all UD layers, including morphological segmentation and features, POS tags and dependency trees, and report first UD Hebrew agreement scores
    \item New SOTA results on the standard benchmark for Hebrew segmentation, tagging and parsing
    \item The first cross-corpus evaluation of out-of-domain (OOD) Hebrew NLP across all UD tasks; we also perform error analysis indicating some issues with previous benchmark data
    \item We release all code and trained models for the tools evaluated in Section \ref{sec:nlp}, including new models for all tasks using the popular Stanza and Trankit libraries, as well as a new SOTA library tailored for Hebrew NLP
\end{enumerate}

\section{Previous work}

In terms of material, there is only one existing TB of modern Hebrew prior to our work \cite{SimaanItaiWinterEtAl2001} based on 1990--91 issues of the newspaper Ha'aretz. However, there are multiple versions of this dataset, leading to some confusion. The original TB contained constituent trees, as well as word-internal segmentation and treebank-specific POS tags. This dataset was converted into dependencies first by \citet{goldberg-elhadad-2009-hebrew} using a custom scheme, and later to an early version of UD by \citet{tsarfaty-2013-unified}. The current UD HTB, using an older version of UD V2 guidelines which became invalid in 2018, is described in \citet{sade-etal-2018-hebrew}, and is used below for evaluation using legacy tokenization. An updated version of this dataset released in this paper is based on the 2018 version.

This work represents the second treebank ever produced for modern Hebrew, and the first cross-domain NLP evaluation of dependency parsing for the language. We briefly survey the state-of-the-art in UD Hebrew NLP in Section \ref{sec:nlp}.

\section{The new corpus}

\subsection{Contents}\label{sec:contents}

Our new corpus, \corpname{},\footnote{\url{https://github.com/universalDependencies/UD_Hebrew-IAHLTWiki}} created by the non-profit Israeli Association of Human Language Technology (IAHLT), contains 5K sentences taken from Wikipedia, which were annotated by a team of 6 annotators, who all have either an undergraduate or graduate degree in Linguistics and a robust knowledge of Hebrew syntax, over 6 months using the Grew-Arborator tool \cite{guibon-etal-2020-collaborative}. During annotator training, a member of the UD core group was consulted whenever questions arose in a series of meetings. Data was sampled from 7 topic categories in order to increase lexical and structural diversity: biographies, events, places, health, legal, finance and miscellaneous (see Table \ref{tab:domains}). Inclusion of the last category is intended to introduce some random topic variation into the data, while the former categories were selected in consultation with a consortium of Israeli industry partners based on high interest applications for information extraction regarding people, places, healthcare, etc.

\begin{table}[htb]
\begin{tabular}{lcrr}
\textbf{domain}  & \textbf{documents} & \textbf{tokens} & \textbf{sentences} \\
\hline
\textit{bio}     & 5             & 21,963        & 754            \\
\textit{event}   & 4             & 16,202        & 580   \\
\textit{finance} & 5             & 8,723         & 299   \\
\textit{health}  & 8             & 20,927        & 824            \\
\textit{law}     & 6             & 22,916        & 788            \\
\textit{place}   & 5             & 22,323        & 829            \\
\textit{misc}    & 6             & 27,895        & 965            \\
\hline
\textbf{total}   & 39            & 140,949       & 5,039          \\
        
\end{tabular}
\vspace{-5pt}
\caption{Domains in the \corpname{} corpus.}\label{tab:domains}
\vspace{-10pt}
\end{table}

Rather than sampling random sentences, full Wikipedia entries were selected in order to to allow for future annotation projects at the document level, such as salient entities, coreference resolution, or other types discourse annotation. Domain data varies somewhat in size, though most domains cover roughly 20-30K tokens, with the exception of \textit{finance} and \textit{event}, to which we devoted less space based on our priorities. Together they were given the same total amount of data as a single domain. 

\begin{table*}[bth]
\resizebox{\textwidth}{!}{
\begin{tabular}{llrrr|llrrr}
\multicolumn{5}{c|}{\textbf{Under-represented in HTB}}  & \multicolumn{5}{c}{\textbf{Under-represented in Wikipedia data}} \\
\textbf{word} & \textbf{translation} & \textbf{Wiki} & \textbf{HTB} & \textbf{ratio}          & \textbf{word} & \textbf{translation}          & \textbf{Wiki} & \textbf{HTB} & \textbf{ratio}                         \\
\hline

\begin{cjhebrew}Mynyly.synp|\end{cjhebrew}    & \textit{Penicillins} & 33               & 0      & \cellcolor[HTML]{305496}0        & \begin{cjhebrew}twryyt\end{cjhebrew}        & \textit{tourism}              & 0                & 31              & \multicolumn{1}{r}{\cellcolor[HTML]{C00000}inf} \\
2018          & \textit{2018}        & 22               & 0               & \cellcolor[HTML]{305496}0        & \begin{cjhebrew}Mw/sl/s\end{cjhebrew}         & \textit{day before yesterday} & 0                & 29              & \multicolumn{1}{r}{\cellcolor[HTML]{C00000}inf} \\
\begin{cjhebrew}b\end{cjhebrew}"\begin{cjhebrew}.thl\end{cjhebrew}         & \textit{LGBT}        & 3                & 0      & \cellcolor[HTML]{305496}0        & \begin{cjhebrew}tyywwk|\end{cjhebrew}        & \textit{Kuwait}               & 0                & 26              & \multicolumn{1}{r}{\cellcolor[HTML]{C00000}inf} \\
\begin{cjhebrew}htyyh\end{cjhebrew}         & \textit{was (3.sg.F)}     & 141              & 1               & \cellcolor[HTML]{8EA9DB}0.007 & \begin{cjhebrew}`wb/s\end{cjhebrew}          & \textit{week}                 & 8                & 110             & \cellcolor[HTML]{D70000}13.75                   \\
\begin{cjhebrew}Mwbl'\end{cjhebrew}        & \textit{album}       & 183              & 3               & \cellcolor[HTML]{8EA9DB}0.016 & \begin{cjhebrew}q.h/sm|\end{cjhebrew}          & \textit{game}                 & 9                & 115             & \cellcolor[HTML]{D70000}12.777                \\
\begin{cjhebrew}Myry/s\end{cjhebrew}         & \textit{songs}       & 127              & 6               & \cellcolor[HTML]{B4C6E7}0.047 & \begin{cjhebrew}lb'\end{cjhebrew}           & \textit{but}                  & 24               & 207             & \cellcolor[HTML]{FF4747}8.625                   \\
\begin{cjhebrew}hl.hm|\end{cjhebrew}                   & \textit{disease}     & 35               & 3               & \cellcolor[HTML]{B4C6E7}0.085 & \begin{cjhebrew}hr.t/sm|\end{cjhebrew}         & \textit{police}               & 20               & 149             & \cellcolor[HTML]{FF6565}7.45                    \\
\begin{cjhebrew}tybr`\end{cjhebrew}                  & \textit{Arabic}      & 30               & 4               & \cellcolor[HTML]{D9E1F2}0.133 & \begin{cjhebrew}k|\end{cjhebrew}"\begin{cjhebrew}.h\end{cjhebrew}           & \textit{member of parliament}                   & 7                & 49              & \cellcolor[HTML]{FF6565}7                       \\
\begin{cjhebrew}tynwwy\end{cjhebrew}                 & \textit{Greek}       & 22               & 3               & \cellcolor[HTML]{D9E1F2}0.136 & \begin{cjhebrew}rlwd\end{cjhebrew}          & \textit{dollar}               & 13               & 89              & \cellcolor[HTML]{FEA466}6.846                \\
\begin{cjhebrew}y'r/s'\end{cjhebrew}                  & \textit{credit}      & 11               & 2               & \cellcolor[HTML]{DDEBF7}0.181 & \begin{cjhebrew}tw.s`wm|\end{cjhebrew}        & \textit{Soviet}               & 10               & 37              & \cellcolor[HTML]{FFDBAF}3.7                    
\end{tabular}}
\caption{Words with striking frequency differences sorted by ratio of frequency in HTB vs. the Wiki data. Ratio for words overrepresented in Wikipedia is shaded blue, and for HTB in red.}\label{tab:vocab}
\vspace{-10pt}
\end{table*}

To illustrate some of the lexical differences between the older HTB news data and \corpname{}, Table \ref{tab:vocab} shows words which are over/under represented in each dataset compared to the other (ratio skewed towards wiki in shades of blue, or HTB in red), taken from the top 50 words sorted by the frequency ratio, and constrained to include a maximum of 3 words not appearing in the other dataset. Some of the items on the left relate to period: `2018' is obviously not discussed in HTB, but neither is `LGBT', which was a very new term in 1990 and not yet translated to Hebrew. Other items relate to domain: inclusion of medical articles leads to 33 occurrences of `Penicillin' (0 in HTB) and 10 times more mentions of `disease'. Other differences are stylistic: the spelling \begin{cjhebrew}htyyh\end{cjhebrew}  for the 3rd person singular feminine \textit{hayta} `was' is very common colloquially, but `sub-standard' for Ha'aretz, which always spells it with one `y': \begin{cjhebrew}htyh\end{cjhebrew}.

On the opposite side of the table, we also see period effects: `Soviet' or `Kuwait', which played important roles in world news in 1990; but also common items missing from Wikipedia, such as `day before yesterday', a relative time term unlikely to appear in encyclopedic text, except perhaps in a quotation. Other underrepresented items interact with genre, such as `week' (very common in narratives, less in exposition) or `but' (more typical of argumentative writing).

Overall our corpus contains around 14K lexical items, nearly 7K of which are missing from HTB, not only due to its topics, but also possibly due to the fact that Wikipedia is authored by diverse volunteers, and is not guided by a newspaper's stringent style policy and proofreaders. Sentences in the corpus also trend longer, which may be of value for parser performance on longer sentences: \emph{M} = 25.02 and \emph{SD} = 14.27 tokens in HTB, compared to \emph{M} = 27.97 and \emph{SD} = 15.79 in \corpname{}. 

\subsection{Changes from HTB}\label{sec:changes-htb}

The previous version of UD HTB has been invalid based on the official UD validation page since 2018, meaning several schema changes were needed to conform to the latest UD standards. To make comparison of the datasets and joint training possible, we release and evaluate a newly revised version of UD HTB which is valid by UD 2.10 release standards. This new version was initially created via automatic scripts using the DepEdit Python library \cite{PengZeldes2020} followed by manual post editing and validation, and nearly doubles the total amount of UD data for Hebrew, (over 11K sentences), allowing for cross-domain experimentation. 

\paragraph{Tokenization} Besides correcting thousands of HTB errors and improving consistency, we introduce a major change to tokenization, which ensures that multiword tokens always correspond to the concatenation of their sub-tokens. In particular, we remove inserted possessive \begin{cjhebrew}l/s\end{cjhebrew} $[$šel$]$ `of' between nouns and their citic possessives (as in \ref{ex:shel} for the MWT \begin{cjhebrew}wtyb\end{cjhebrew} $[$beit-o$]$ `his house'), object marking \begin{cjhebrew}t'\end{cjhebrew} $[$et$]$ (accusative marker, as in \ref{ex:et} for \begin{cjhebrew}hyty'r\end{cjhebrew} $[$re'iti-ha$]$ `(I) saw her') and orthographically unexpressed articles (as in \ref{ex:ha} for \begin{cjhebrew}tybb\end{cjhebrew} $[$ba-bait$]$ `in the house').

\ex. \label{ex:shel} \a. Old: \begin{cjhebrew}'wh\end{cjhebrew}-\begin{cjhebrew}$\_$l/s$\_$\end{cjhebrew}-\begin{cjhebrew}tyb\end{cjhebrew} $[$bait \_shel\_ hu$]$
    \b. New: \begin{cjhebrew}w\end{cjhebrew}-\begin{cjhebrew}tyb\end{cjhebrew} $[$beit o$]$ 
            
\ex. \label{ex:et} \a. Old: \begin{cjhebrew}'yh\end{cjhebrew}-\begin{cjhebrew}$\_$t'$\_$\end{cjhebrew}-\begin{cjhebrew}yty'r\end{cjhebrew} $[$ra'iti \_et\_ hi$]$
    \b. New: \begin{cjhebrew}h\end{cjhebrew}-\begin{cjhebrew}yty'r\end{cjhebrew} $[$re'iti ha$]$ 

\ex. \label{ex:ha} \a. Old: \begin{cjhebrew}tyb\end{cjhebrew}-\begin{cjhebrew}$\_$h$\_$\end{cjhebrew}-\begin{cjhebrew}b\end{cjhebrew} $[$be \_ha\_ bait$]$
    \b. New: \begin{cjhebrew}tyb\end{cjhebrew}-\begin{cjhebrew}b\end{cjhebrew} $[$ba bait$]$

In \ref{ex:shel}, the former HTB tokenization inserted the word \textit{shel} `of' surrounded by underscores before clitic possessors, as though the example reads `house\_of\_him', rather than `his house'. This complicates tokenization and introduces an unnecessary inconsistency with related languages in UD, such as UD Arabic, which does not insert an unexpressed preposition in the same construction, nor object markers in cases like \ref{ex:et}. 

The last case in \ref{ex:ha}, i.e.~removal of zero articles (which speakers can reconstruct from context while reading Hebrew but are not trivial to predict) is the only change resulting in a loss of information, and we therefore replace these by a standard UD morphological feature \texttt{Definite=Def}. As a result, conversion between the new and old TB tokenization is deterministic in both directions, but the new tokenization style is both simpler and matches practices in related languages, most notably Arabic.

\paragraph{Dependencies} Some custom relation sub-types, which were predictable from the words they connect, have been removed in the new scheme, such as \texttt{case:acc} and \texttt{case:gen} for object and possessive markers, or \texttt{mark:q} for the question markers.

\subsection{Validation}

\begin{figure*}[t!bh]
\centering
\includegraphics[width=\textwidth]{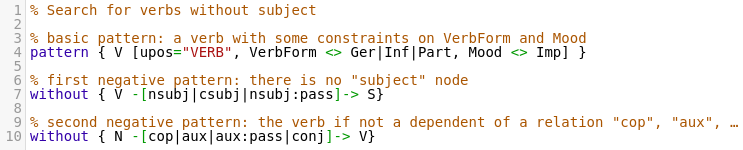}
\includegraphics[clip,width=0.7\textwidth,trim = 0cm 0cm 0cm 0cm]{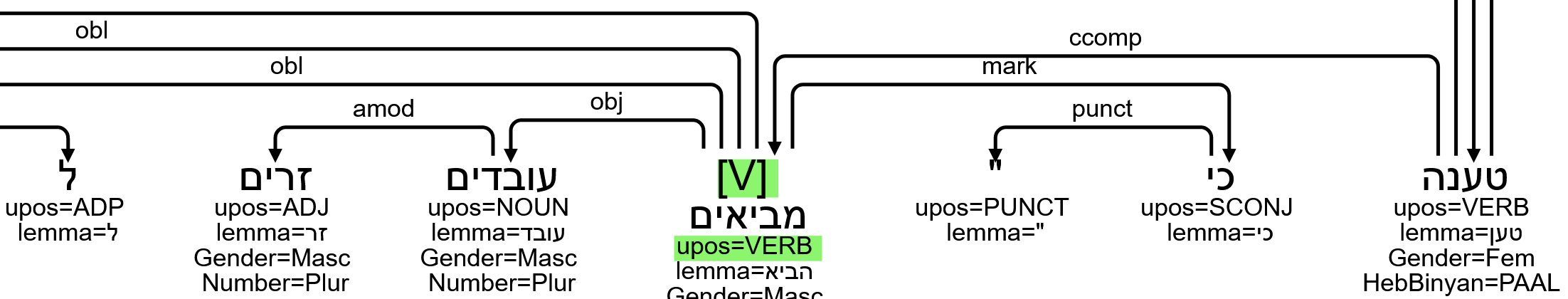}
\caption{A Grew search in HTB for subjectless verbs and a result.}\label{fig:grew}
\vspace{-12pt}
\end{figure*}

The standard UD toolkit\footnote{\url{http://github.com/UniversalDependencies/tools}}
contains a validation framework, but at the language-specific level it
is quite limited, largely checking feature-POS combination sanity and permitted
relations or auxiliaries. Ideally, each enforceable provision in language specific guidelines should be captured in machine-readable format, a vision we attempt to implement using \texttt{grewv}, an extension of the Grew ``graph rewriting for NLP'', a
search and transformation engine for decorated graphs, originally
designed for corpus exploration \cite{Guillaume2021}. Grew allows rigorous definition of graph patterns, incl. quantification and negation, which enables e.g.~searching for verbs with no subject (Figure \ref{fig:grew}). We created a tool around grew, dubbed
\texttt{grewv} (for ``grew validator''), which uses
grew expressions to describe \emph{non-conformant} trees and generate
an error report in human- and machine-readable
formats.

As we encountered new phenomena in the data, annotators worked together to define guidelines, but also rules, consisting of a graph matching definition, a short name, a long-form message template, and an error level: `error' or `warning'. Rule templates
support placeholders for tree-specific
information, such as token numbers. While errors result
in validation failure, warnings can be dismissed, indicating that an annotator has reviewed and approved the data. An error-level message and pattern are shown in \ref{ex:rule}, which prohibits tokens from bearing the \texttt{cc} label for coordinating conjunctions if they are not the child of a coordinated token, with one Hebrew lexical exception for the expression \begin{cjhebrew}Nybw Nyb\end{cjhebrew} `both ... and'.

\vspace{-5pt}

\ex. \textbf{error:} \textit{Token} \{matching[nodes][X]\} \textit{has a cc child} (token \{matching[nodes][Y]\})\textit{, but is neither conj, parataxis nor root.} \\
  \textbf{pattern:} \{Y[lemma<>"\begin{cjhebrew}Nyb\end{cjhebrew}"]; X-[cc]->Y;\} \\without \{ * -[conj|root|parataxis]-> X;  \}	\label{ex:rule}

\vspace{-5pt}

We integrated \texttt{grewv} into all stages of the
project: continuous validation is used on the ruleset and at
annotation time, and the treebank is passively scanned for trees
requiring correction on updates. Changes to the ruleset are submitted, discussed, and reviewed over a mailing list, and proposed changes trigger an analysis job on our
continuous integration system. Each change is checked for syntax errors
and illustrative examples of trees which are
supposed to pass and fail. Finally, a random subsample of outputs is inspected by an
annotator to confirm that the rule works as expected.

Confirmed changes are automatically deployed to our Arborator-Grew
annotation instances; we have made modifications to track and report
validation errors and warnings, and annotations cannot be marked `final'
until they have passed validation. As rules evolved, the treebank is passively scanned for
`stale' trees: annotations which passed validation at the time they
were made, but now require correction in order to pass the latest
ruleset. Periodically batches of stale trees are prepared for
freshening, to ensure the treebank asymptotically approaches a
consistent annotation level.

\subsection{Inter-annotator agreement (IAA)}

Table~\ref{tbl:iaa} gives aligned token accuracy using the official CoNLL scorer (Words) and Cohen's $\kappa$ for each annotation for $\sim$440 double annotated sentences ($\sim$12.6K tokens) before adjudication and \texttt{grewv} validation. As this is the first Hebrew corpus annotated natively in UD, these are, to the best of our knowledge, the first reported IAA scores on full UD annotation for the language. Due to the nature of $\kappa$, features beyond segmentation are computable only for identically tokenized words, however due to the high segmentation score this is negligible.

\begin{table}[h!tb]
\centering
\resizebox{\columnwidth}{!}{
\begin{tabular}{r|r|r|r|r|r|r}

  \textbf{Words}   
& \textbf{Lemma}   
& \textbf{UPOS}    
& \textbf{FEATS}
& \textbf{Head}    
& \textbf{Deprel}  
& \textbf{Misc}    
\\ \hline
  \(99.1\%\)
& \(98.5\)
& \(94.8\)
& \(90.9\)
& \(95.8\)
& \(95.7\)
& \(95.6\)
\end{tabular}}
\caption{Word segmentation accuracy, and Cohen's kappa for each annotation layer. Misc stands for miscellaneous optional UD annotations, including e.g.~\texttt{CorrectForm} for typos.}
\label{tbl:iaa}
\vspace{-10pt}
\end{table}

The metrics indicate a very high level of agreement, with lowest scores on \textsc{feats}, where agreement holds only if all features agree (Gender, Number, Tense etc.); for $\kappa$-scores on individual morphological features, see the Appendix.

\section{In-domain and cross-corpus NLP}\label{sec:nlp}

In this section we present a series of experiments on cross-corpus segmentation, tagging, lemmatization and parsing, which is made possible for the first time by the release of non-newswire treebank data. Several systems exist for UD parsing of Hebrew: in this section we compare two popular neural off the shelf, end-to-end UD systems: Stanza \cite{QiEtAl2020} and Trankit \cite{NguyenEtAl2021}, the latter of which is the SOTA for Hebrew parsing; and the current SOTA system for non-concatenative Hebrew segmentation and tagging \cite{seker-etal-2022-alephbert}, which is the only one using a pre-trained Hebrew Transformer introduced in the same paper, AlephBERT, but which does not cover UD parsing based on the paper.\footnote{Although the AlephBERT model has been released, code from the paper does not enable re-running training of the NLP components; we therefore report the scores from the paper below.} We release and describe our own system for these tasks, called HebPipe,\footnote{\url{https://github.com/amir-zeldes/HebPipe/}} below, which is a pipeline system building on the best existing systems with some incremental improvements.

\paragraph{Segmentation} All previous work on UD Hebrew parsing has been limited by the use of non-concatenative tokenization schemes. However with the conversion to the concatenative tokenization described in Section \ref{sec:changes-htb}, we are able to take advantage of high accuracy characterwise segmentation approaches using binary classification (each character either begins a new segment or not). The previous characterwise SOTA system, RFTokenizer \cite{Zeldes2018a} is based on lexical features, such as looking up possible POS tags of MWT substrings around a segmentation point, and an XGBoost classifier, and has been applied to other morphologically rich languages, such as Arabic and Coptic \cite{ZeldesSchroeder2016}; however it does not utilize pretrained transformers, which do not generally encode character level information (AlephBERT is word-piece based). \citet[51]{seker-etal-2022-alephbert} also mention that it is counterintuitive to use a word level model as input to character level tasks, but note that despite this their approach still achieves a high segmentation score.

To improve the SOTA in segmentation, we combine the characterwise RFTokenizer classifier with the transformer from \citet{seker-etal-2022-alephbert} operating at the MWT word level. We train an LSTM using AlephBERT to predict \textit{whether} each word contains prefixes requiring segmentation, suffixes, or both, thereby targeting an MWT-level property. We then feed these predictions to RFTokenizer as a feature, 
an approach not previously applied to the problem to the best of our knowledge. 
To avoid overfitting, we reserve half of the training data and the development set for the AlephBERT LSTM training, and the remaining half for the XGBoost classifier, which is fed the features described in \citet{Zeldes2018a} as well as the AlephBERT predictions (code attached to this paper). 

\paragraph{Tagging} For POS tagging and features we employ a simple LSTM sequence tagger using the same pre-trained AlephBERT embeddings. We observe in development data that some tokens are ambiguous regarding part of speech depending on whether they are prefixes or suffixes: for example, the letter vav $\langle$w$\rangle$ can stand for the word `and', pronounced $[$ve$]$ and tagged \textsc{cconj}, at the beginning of a MWT, or the word `him/his', a pronoun pronounced $[$o$]$ at the end of a MWT. If tokens are tagged as a naive sequence, these are occasionally mistagged in ambiguous contexts, for example:

\ex. \label{ex:ve} \quad \begin{cjhebrew}Nwmr' w tyb\end{cjhebrew} \quad $\langle$byt w 'rmwn$\rangle$
    
If the first two tokens are spelled together (\begin{cjhebrew}wtyb\end{cjhebrew}), then the example is a predicative verbless sentence (`his house is a palace'), and the letter `w' is a pronoun (\textsc{pron}). However if the last two tokens are spelled together (\begin{cjhebrew}Nwmr'w\end{cjhebrew}), then `w' is a coordinating conjunction (\textsc{cconj}), meaning `and', and the translation would be `a house and a palace'. Other errors can occur similarly, for example between an initial article \textit{ha-} `the' spelled the same as the suffix feminine possessive \textit{-a} `her'.

To make the model aware of such cases, we concatenate a dense embedding using BIES notation to each token vector, indicating whether it begins a MWT (B), ends one (E), is inside one (I) or represents a single-token word (S). Using concatenative tokenization, we observe a degradation of -0.32\% in end-to-end tagging accuracy on HTB by ablating this feature, which amounts to an error reduction of $\sim$12\%. 
Although this improvement is minor, it is not computationally complex, and we are hopeful that it could be a useful approach for other MRLs, such as Arabic.
Tagger predictions are also fed into Stanza's lemmatizer to produce our lemmatization.

\paragraph{Parsing} For dependency parsing we rely on the prevalent biaffine approach proposed by \cite{DozatManning2017}, using the implementation in DiaParser \cite{attardi-etal-2021-biaffine}, which adds intermediate transformer layer inputs to the transition classifier. We again rely on AlephBERT to represent input embeddings. 

We use the default hyperparameters for all components described above and do not perform hyperparameter optimization (see appendix for details).

\paragraph{In-domain results} Table \ref{tab:results} compares performance for the systems listed above. Results are computed using the official CoNLL scorer from the 2018 shared task on UD parsing, and with the exception of using gold sentence splits to match prior work on Hebrew, 
all numbers reflect realistic, end-to-end performance from plain text input, using the MWT F1 score for morphological segmentation, and AligndAcc for all other metrics. All numbers have been reproduced by retraining each system, except for \citeauthor{seker-etal-2022-alephbert}'s system, for which trainable code is not provided. For HTB we report results using the concatenative tokenization, as well as with the older non-concatenative tokenization (in brackets) for comparability with previous work, with the official train-dev-test splits maintained. The splits for \corpname{} were established by stratified sampling evenly across the domains described in Section \ref{sec:contents}. 

\begin{table*}[tbh]
\resizebox{\textwidth}{!}{
\begin{tabular}{l|llll|lll}
& \multicolumn{4}{c|}{\textbf{train on HTB $\rightarrow$ test on HTB}} & \multicolumn{3}{c}{\textbf{train on Wiki  $\rightarrow$ test on Wiki}} \\
\textit{\textbf{}}      & \textbf{Stanza}     & \textbf{Trankit}    & \textbf{Seker et al.} & \textbf{This paper} & \textbf{Stanza}      & \textbf{Trankit}      & \textbf{This paper}      \\
\hline
MWT F1              & 93.97 (92.82)       & 97.27 (96.04)       & -- (98.20)*                  & \textbf{98.81 (98.37)}       & 92.87                & 94.64                 & \textbf{98.78}                    \\
POS                     & 96.99 (97.12)       & \textbf{97.46 (97.63)}       & -- (96.20)*                  & 97.34 (97.40)        & 95.76                & 96.98                 & \textbf{97.27 }                   \\
FEATS                   & \textbf{95.45 (95.65)}       & 85.95 (85.65)       & -- (93.05)*                 & 91.68 (92.52)       & 89.23                & 90.69                 & \textbf{91.06}                    \\
AllTags                 & \textbf{94.64 (94.85)}       & 85.20 (84.90)         & --                    & 91.06 (91.92)       & 88.24                & 89.58                 & \textbf{90.30}                     \\
Lemma                   & 96.63 (96.89)       & 96.69 (97.06)       & --                    & \textbf{97.52 (97.58)}       & \textbf{97.51}                & 94.70                  & 97.49                    \\
UAS                     & 85.62 (85.78)       & 85.46 (85.60)        & --                    & \textbf{91.90 (92.07)}        & 82.96                & 89.81                 & \textbf{92.19}                    \\
LAS                     & 82.67 (82.88)       & 82.82 (83.01)       & --                    & \textbf{89.42 (89.65)}       & 80.22                & 87.65                 & \textbf{90.01}                    \\
CLAS                    & 76.01 (75.93)       & 83.96 (83.88)       & --                    & \textbf{84.48 (84.48)}       & 73.16                & 83.96                 & \textbf{86.16}                    \\
MLAS                    & 70.63 (68.97)       & 63.00 (60.20)           & --                    & \textbf{72.24 (72.24)}       & 59.64                & 69.35                 & \textbf{72.37}                    \\
BLEX                    & 72.71 (72.65)       & 79.80 (79.87)        & --                    & \textbf{80.99 (80.99)}       & 70.66                & 76.03                 & \textbf{82.56}                   
\end{tabular}}
\caption{In-domain UD NLP performance for both datasets. Figures in brackets are for the older, non-concatenative tokenization. All numbers are end-to-end from plain text. A `*' indicates numbers from the cited paper, other numbers are reproduced by the authors.}\label{tab:results}
\vspace{-10pt}
\end{table*}

As the table shows, our approach achieves new SOTA scores for segmentation (MWT F1), lemmatization, parsing (UAS and LAS) as well as morphologically and lexically informed combined UD scores (CLAS, MLAS, BLEX) on HTB.\footnote{See \url{https://universaldependencies.org/conll18/evaluation.html} for details on these metrics} For tagging, the result is very close -- given that both Trankit and our system use a torch-based, transformer-driven sequence tagger, it appears that either Trankit's underlying XLM-RoBERTa model \cite{conneau-etal-2020-xlmroberta} is superior to AlephBERT here, or we are looking at minor, random chance differences. Overall, the most notable advances in score are for segmentation -- likely due to stacking RFTokenizer and AlephBERT, since the next best score is Seker et al.'s pure AlephBERT system -- and for parsing, likely due to the use of a language specific BERT, which has not been reported on in previous papers for Hebrew. Here the difference to Trankit's XLM-RoBERTa result is very noticeable (+6.6 points), however we note that this folds in gains from the superior segmentation quality, since all scores are end-to-end from plain text sentences.

For \corpname{} the results favor our approach even more, with the sole exception of lemmatization. However here our approach in fact uses Stanza itself, meaning the slightly lower score is probably due to random initialization differences. Although the absolute numbers for \corpname{} are sometimes lower than for HTB's scores, we note that the Wiki data may not only be more challenging, but the use of the simplified tokenization shifts task difficulties: based on HTB, segmentation accuracy is higher across the board when using the simpler tokenization; however due to the need to represent orthographically unexpressed articles using morphological features and the lack of easy-to-tag inserted pseudo-tokens, metrics involving morphology may become lower.

\paragraph{Cross-domain results} In order to assess how reliably our results generalize to unseen data from a different source, we test the trained models from Table \ref{tab:results} on data from the `other' corpus, using the same partitions (training still relies on each source corpus `train' partition, and evaluation is on the same `test' partition used in the previous table). For this cross-corpus evaluation, we report scores only on the new tokenization, since no previous scores are available for comparison for this setting, and systems are unable to learn non-concatenative MWT expansion from the new Wiki dataset. Table \ref{tab:cross-corpus} shows the results for Stanza, Trankit and our system, HebPipe.

\begin{table*}[tbh]
\centering
\begin{tabular}{l|rrrr|rrrr}
        & \multicolumn{4}{c|}{\textbf{train on Wiki $\rightarrow$ test on HTB}}                                            & \multicolumn{4}{c}{\textbf{train on HTB $\rightarrow$ test on Wiki}}                  \\
        & \textbf{Stanza}              & \textbf{Trankit}             & \textbf{This paper} & \textbf{$\Delta$}          & \textbf{Stanza} & \textbf{Trankit} & \textbf{This paper} & \textbf{$\Delta$}         \\
        \hline
MWT F1  & 91.79 & 92.24 & \textbf{98.59}               & {\color[HTML]{FE0000} -0.22}  & 91.79           & 93.03            & \textbf{98.51}               & {\color[HTML]{FE0000} -0.27} \\
POS     & 94.09 & 94.27 & \textbf{95.25}               & {\color[HTML]{FE0000} -2.09}  & 94.09           & 95.02            & \textbf{96.41}               & {\color[HTML]{FE0000} -0.86} \\
FEATS   & 83.35 & 81.19 & \textbf{88.02}               & {\color[HTML]{FE0000} -3.66}  & 83.35           & 76.50             & \textbf{90.06}               & {\color[HTML]{FE0000} -1.00}    \\
AllTags & 82.12 & 79.72 & \textbf{86.97}               & {\color[HTML]{FE0000} -4.09}  & 82.12           & 75.10             & \textbf{88.94}               & {\color[HTML]{FE0000} -1.36} \\
Lemma   & 95.05 & 90.96 & \textbf{97.31}               & {\color[HTML]{FE0000} -0.21}  & 95.05           & 96.41            & \textbf{97.75}               & {\color[HTML]{009901} 0.26}  \\
UAS     & 76.50  & 82.55 & \textbf{86.05}               & {\color[HTML]{FE0000} -5.85}  & 76.50            & 82.32            & \textbf{90.37}               & {\color[HTML]{FE0000} -1.82} \\
LAS     & 72.26 & 78.22 & \textbf{82.14}               & {\color[HTML]{FE0000} -7.28}  & 72.26           & 78.34            & \textbf{87.35}               & {\color[HTML]{FE0000} -2.66} \\
CLAS    & 63.65 & 72.62 & \textbf{76.29}               & {\color[HTML]{FE0000} -8.19}  & 63.65           & 77.60             & \textbf{83.02}               & {\color[HTML]{FE0000} -3.14} \\
MLAS    & 45.19 & 47.25 & \textbf{60.22}               & {\color[HTML]{FE0000} -12.02} & 45.19           & 45.10             & \textbf{67.90}                & {\color[HTML]{FE0000} -4.47} \\
BLEX    & 58.63                        & 60.88                        & \textbf{72.86}               & {\color[HTML]{FE0000} -8.13}  & 58.63           & 72.88            & \textbf{79.73}               & {\color[HTML]{FE0000} -2.83}
\end{tabular}
\caption{Out-of-domain UD NLP performance, training on one corpus and testing on the other, all numbers end-to-end from plain text. $\Delta$ gives the difference between HebPipe in-domain and out-of-domain performance.}\label{tab:cross-corpus}
\vspace{-10pt}
\end{table*}

As the table shows, HebPipe achieves the best cross-corpus generalization, primarily due to large gains in MWT F1, which reduce cascading errors in downstream tasks. Segmentation degradation is only around 0.25\%, meaning POS tagging and lemmatization are mainly vulnerable to OOV items, which are mitigated by AlephBERT's lexical coverage. Stanza and Trankit both suffer only a little over 1--2\% tagging degradation, but for parsing metrics degradation is much more substantial (around 5--10\% LAS on HTB). 

Across the board, degradation is more substantial when predicting on HTB and especially on dependency metrics, which suggests that HTB contains more constructions not represented in the Wiki data. However, our error analysis suggests that a substantial portion of parsing degradation is owing to errors in the original HTB's conversion into UD (see Section \ref{sec:err} below). Overall, these results indicate that OOD performance on a second genre of formal, written Hebrew is fairly reliable for segmentation and tagging, but less so for parsing, and that the segmentation approach taken in this paper is particularly robust, possibly due to XGBoost's well known resistance to overfitting. We stress however that these results do not yet indicate performance quality on informal text types.

\paragraph{Joint training} In response to reviewer feedback in the rebuttal period, we were able to train joint models for each component, using both the HTB and \corpname{} train set for training and the joint dev sets for early stopping via simple concatenation. Table \ref{tab:joint} gives the scores for the jointly trained model. While we do not necessarily recommend using a joint model due to possible inconsistencies between the datasets which have not yet been resolved (see Section \ref{sec:err} below), the numbers indicate that the model is able to robustly deal with both test sets, with neither substantial gains nor degradation over in-domain training numbers. Importantly, the model clearly outperforms cross-domain results, suggesting that it may indeed be overall the most robust choice for totally unseen data. We expect that further harmonization of the two datasets will increase the usefulness of joint training.

\begin{table}[h!bt]
\resizebox{\columnwidth}{!}{
\begin{tabular}{l|ll||l|ll}
        & \textbf{HTB} & \textbf{Wiki} &      & \textbf{HTB} & \textbf{Wiki} \\
        \hline
MWT F1  & 98.78        & 98.65         & UAS  & 91.11        & 92.65         \\
POS     & 97.14        & 97.29         & LAS  & 88.29        & 90.36         \\
FEATS   & 91.02        & 90.93         & CLAS & 83.27        & 86.65         \\
AllTags & 90.39        & 90.11         & MLAS & 70.57        & 72.37         \\
Lemmas  & 97.65        & 98.15         & BLEX & 80.13        & 84.03        

\end{tabular}
}
\caption{UD NLP performance for a joint model, trained on both corpora.}\label{tab:joint}
\vspace{-5pt}
\end{table}

\section{Error analysis}\label{sec:err}

The discrepancy in NLP quality across corpora for different tasks, and especially lower OOD scores below the segmentation level, leads us to suspect that while segmentation in the two corpora is largely mutually consistent (after conversion of the tokenization scheme), other annotation layers may have some issues. Such issues could also explain why joint training does not outperform in-domain scores -- improvements from having more training data could be cancelled out by inconsistencies. In order to better understand the most common errors, Table \ref{tab:errs} gives the three most commonly confused label pairs for POS tags and dependency relations in both cross-corpus directions (for complete confusion matrices, see the Appendix).

\begin{table}[h!tb]
\resizebox{\columnwidth}{!}{
\begin{tabular}{l|lll|lll}
                & \multicolumn{3}{c|}{\textbf{train Wiki$\rightarrow$ test HTB}} & \multicolumn{3}{c}{\textbf{train HTB$\rightarrow$ test Wiki}} \\
                & \textbf{gold}   & \textbf{pred}   & \textbf{freq}   & \textbf{gold}   & \textbf{pred}   & \textbf{freq}   \\
                \hline
\textit{POS}    & NOUN            & PROPN           & 110             & PROPN           & NOUN            & 160             \\
\textit{}       & VERB            & ADJ             & 86              & VERB            & AUX             & 23              \\
\textit{}       & NOUN            & PROPN           & 73              & ADJ             & VERB            & 21              \\
\hline
\textit{deprel} & nmod            & obl             & 77              & obl             & nmod            & 73              \\
                & compound        & nmod            & 70              & nmod            & obl             & 72              \\
                & obl             & nmod            & 60              & compound        & flat            & 30             
\end{tabular}
}
\caption{Top 3 confused tag and deprel pairs by corpus.}\label{tab:errs}
\vspace{-10pt}
\end{table}

As the table shows, a substantial portion of errors is due to NOUN vs.~PROPN confusions, which is not unusual in general (see \citealt{BehzadZeldes2020}) but suspicious given that directions are flipped: testing on HTB we see over-prediction of proper nouns, while testing on Wikipedia data shows the opposite trend. Manual inspection of the errors reveals that HTB's name annotations are inconsistent: names in the original UD HTB data, and especially place names, are tagged NOUN if they are transparently composed of multiple Hebrew nouns. Thus places like \textit{Tel Aviv} (lit. `Spring Mound') or \textit{Kfar Saba} (lit. `Grandfather Village') are not tagged as PROPN in HTB, but are PROPN in \corpname{}, leading to prediction errors in both directions. ADJ vs. VERB confusions are also `flipped', though less common, and are owing to a number of lexical items with different treatments, such as \textit{dome} `similar' and \textit{shone} `different', which are tagged as adjectives in the Wiki data, but as participles in HTB, based on their etymology.

For dependency relations, most confusions are due to \texttt{obl} vs. \texttt{nmod}, which are annotated consistently in the data and correspond to verbal and nominal prepositional modifiers, i.e.~the well known challenge of PP attachment (see \citealt{KawaharaKurohashi2005}). Manual inspection confirms that these are less worrying in terms of corpus compatibility. Confusions involving the \texttt{compound} label are mainly due to errors in HTB, which include common items such as \textit{brit ha-mo'atsot} `Soviet Union' (occasionally labeled \texttt{nmod} instead of \texttt{compound}), and often carry an additional annotation \texttt{ConvUncertainLabel} in the original data, indicating that this is an issue with the original conversion of the data to UD. We plan to address these and other inconsistencies in future releases of the data.

\section{Conclusion}

In this paper we make available, describe and evaluate annotation quality in the first non-newswire treebank for Hebrew, complementing the existing newswire data which is by now over 30 years old. We propose and implement a new, simpler tokenization scheme for Hebrew, and release a revised version of the existing UD Hebrew newswire corpus which follows the new scheme. Our evaluation of NLP systems on the new tokenization indicate that it is somewhat easier for a range of tokenizers, at the cost of slightly lower scores for downstream morphosyntactic annotations. We also evaluate agreement on the complete UD annotation pipeline for Hebrew for over 12K doubly annotated tokens, and build a novel infrastructure for error detection based on Grew.

As part of our evaluation of NLP quality on the new resources, we compared several off the shelf systems, whose approaches we combine and incrementally improve to produce a new system, HebPipe, which is released with this paper. Our system achieves new SOTA results on most UD NLP tasks for the older HTB dataset, including segmentation, lemmatization and dependency parsing, evaluated on both the new and older tokenization schemes. We also reported on the first cross-corpus parsing experiment results, which indicated that segmentation and tagging using our system generalize well and substantially better than previous systems, while full dependency parsing is still less reliable out of domain. Error analysis indicates that some of the issues are caused by remaining compatibility problems across the corpora, which we plan to address in the future.

In other future work, we are in the process of annotating data from further sources, including written language from blogs, social media and government websites, as well as spoken data from parliament proceedings and TV shows. To improve annotation, we hope to extend \texttt{grewv} to annotator-mediated autocorrection of validation errors including choosing between multiple correction options, as well as machine-guided refinement of annotation guidelines, and guiding source selection for collecting new material. 

URLs for \corpname{}, the revised UD HTB and complete code for the NLP tools from Section \ref{sec:nlp} will be included after review of this paper.

\section*{Limitations}

While the resources created for this paper substantially broaden the genres, domains and temporal diversity of data for Hebrew segmentation, tagging and parsing, they are still very limited. In particular, the newly released data reflects relatively formal and well edited language, and does not cover spoken language or informal user-generated content from the web, areas which we would like to explore in future work. Statements about generalizability and OOD reliability of tools should therefore be interpreted cautiously. Our results are only truly reliable for the specific case of the Hebrew datasets examined here, and may not apply to other domains or similar morphologically rich languages (MRLs), such as Arabic, with which we are also experimenting.

The NLP tools compared in this paper also constitute a narrow and subjective selection -- the choice of Stanza and Trankit was motivated by their popularity and the existence of previous models for Hebrew, while comparison with \citeauthor{seker-etal-2022-alephbert}'s work was due to the paper's previous SOTA scores. It is possible that other architectures could outperform the results reported here, as well as reach different conclusions about generalizability and error sources. Due to the large number of pipeline components in the systems compared here, each with separate trained models for each corpus (18 models per experiment across 3 systems), and the focus of this paper on the corpus resource, we decided to only use seeds and not to obtain average scores from a large number of runs, which would require spending substantially more computing resources and promote current carbon-intensive trends in NLP. While we understand the desire for such numbers in papers focused on novel parsing architectures with very close numbers, we feel that they would not change the results presented here (though reviewers can let us know otherwise).

For similar reasons, we performed no hyperparameter optimization on any of the systems, meaning that it is also possible that systems would fare differently if this were attempted, possibly making some of our incremental improvements, such as within-MWT BIES positional embeddings, unhelpful. That said, we are skeptical that extensive optimization against single-genre dev sets is meaningfully useful for performance on unseen data in the wild, and we prefer to invest in developing systems which can be trained and run with moderate or no GPU resources (see also the Ethics Statement below). Optimizing pipeline architectures thoroughly would mean considering interactions between different components' models, for gains that would quite possibly be limited to the data used in the paper itself.

Finally, the evaluation of our corpus focused on end-to-end numbers, rather than evaluating each task on gold inputs (i.e.~parsing numbers reflect upstream tokenization errors, etc.), but did use gold sentence segmentation, to match the setup in previous SOTA work such as \citet{seker-etal-2022-alephbert}. Especially given the limited space in this paper, we aimed to follow the UD shared task paradigm in preferring end-to-end scores, which more closely mirror expected performance in the wild. Although this complicates the interpretation of better and worse downstream component choices, we feel this is inevitable, since upstream outputs such as segmentation and tagging are also part of the input for downstream tasks, and a large number of evaluation scenarios is conceivable. By contrast, our inter-annotator agreement study required identical tokenization to compute kappa for annotations, meaning the lack of agreement numbers including tokenization disagreements is a further limitation of our results. This last issue likely has low impact on numbers due to the high agreement on tokenization, and we intend to release the doubly annotated raw data for interested researchers as well. 

\section*{Ethics Statement}\label{ethics}

This work contributes to open source and open access progress in NLP for morphologically rich languages, and specifically for Hebrew, which has not enjoyed the same wealth of resources as English and other European languages. We recognize that NLP research has a computing cost and carbon footprint, which motivates us to release all models in this work (preventing the need to retrain similar models), to use base-sized language models, and to avoid extensive hyperparameter optimization on these single-genre datasets, which may lead to minor improvements on test sets, but may or may not generalize to applications in the wild. 

We also recognize that NLP tools can be used to do harm, but expect that the type of NLP processing promoted here will do more good than harm by preventing tools from adhering closely to outdated and narrow-domain data, which this work aims to broaden. Given that systems for UD-style outputs for Hebrew already exist, we view any reduction in topical and authorial bias, as well as the public release of more resources, as net positives. All participants in this work have been compensated. The annotators (3 female, 3 male) were employed as regular employees of IAHLT, the non-profit organization which funded the treebank. Previous work has been credited to the best of our knowledge.

\section*{Acknowledgements}

We are grateful for the kind support of the Israel Innovation Authority and the Israel National Digital Agency. We would also like to acknowledge the immense help of the Israeli Association of Human Language Technologies (\url{iahlt.org}), its members and employees for supporting this work, as well as all of the annotators who participated in this project, without whose hard work this resource would not have been possible. 

\bibliography{anthology,custom}
\bibliographystyle{acl_natbib}

\appendix

\section{Agreement on morphological features}

Table~\ref{tbl:iaa-feat} details kappa scores for each morphological feature category exhibiting more than 10 disagreements, including both universal UD features, and Hebrew-specific ones, such HebBinyan (the morphological inflectional class of Hebrew verbs). The IAA study was performed before the introduction of the \texttt{grewv} validation system, and running the system on the double annotated data reveals that in over 29\% of the disagreements (387 of 1,297 cases) at least one of the annotators' decision would have been flagged as an error, based on the rules in Table~\ref{tbl:iaa-grewv}. This result underscores the importance of automatic validation, and the possible lesson that future IAA studies should report the impact of validators on potential disagreement errors that would be prevented by validation.

\begin{table}[h!tb]
\centering
{\small %
\begin{tabular}{l|rr}
   \textbf{Label} & \textbf{Kappa} & \textbf{Disagreements}
\\ \hline
Case             & 0.902          & 82               \\
Definite         & 0.966          & 123              \\
Foreign          & 0.434          & 13               \\
Gender           & 0.954          & 281              \\
HebBinyan        & 0.960          & 56               \\
Number           & 0.959          & 238              \\
NumType          & 0.688          & 48               \\
Person           & 0.966          & 65               \\
Polarity         & 0.709          & 48               \\
Prefix           & 0.592           & 11               \\
PronType         & 0.976          & 68               \\
Tense            & 0.968        & 42               \\
Typo             & 0.482          & 15               \\
VerbForm         & 0.937          & 37               \\
VerbType         & 0.703          & 31               \\
Voice            & 0.900          & 139              \\
\hline
\textit{average} & \textbf{0.818} & \textbf{81.06} \\
\textit{total}   &                   & \textbf{1297}   
\end{tabular}
}
\caption{Cohen's Kappa and the total number of disagreements for each
feature and miscellaneous label.}
\label{tbl:iaa-feat}
\end{table}

\begin{table}[h!tb]
\centering
\resizebox{\columnwidth}{!}{
\begin{tabular}{llr}
   \textbf{Rule name} & \textbf{Feature} & \textbf{Differences}
\\ \hline
    \texttt{verb-mand-voice}         &  \texttt{Voice}            &  \(42\)
\\ \texttt{pass-arg}                &  \texttt{Voice}            &  \(32\)
\\ \texttt{verb-mand-binyan}        &  \texttt{HebBinyan}        &  \(28\)
\\ \texttt{def-cons}                &  \texttt{Definite}         &  \(25\)
\\ \texttt{binyan-nomid}            &  \texttt{Voice,HebBinyan}  &  \(22\)
\\ \texttt{pron-prontype-required}  &  \texttt{PronType}         &  \(18\)
\\ \texttt{aux-nopolarity}          &  \texttt{Polarity}         &  \(17\)
\\ \texttt{amod-onlyadj}            &  --                         &  \(15\)
\\ \texttt{aux-binyan}              &  \texttt{HebBinyan}        &  \(13\)
\\ \texttt{adp-case-gen}            &  \texttt{Case}             &  \(13\)
\\ \texttt{yesh-polarity}           &  \texttt{Polarity}         &  \(12\)
\\ \texttt{pron-nopolarity}         &  \texttt{Polarity}         &  \(12\)
\\ \texttt{noun-adj-gender-agr}     &  \texttt{Gender}           &  \(11\)
\\ \texttt{nmod-obl-case}           &  \texttt{Case}             &  \(11\)
\\ \texttt{mandatory-gender}        &  \texttt{Gender}           &  \(11\)
\\ \texttt{cc-child-conj}           &  --                        &  \(11\)
\\ \texttt{noun-adj-num-agr}        &  \texttt{Number}           &  \(10\)
\\ \texttt{mandatory-number}        &  \texttt{Number}           &  \(10\)
\\ (other types)                    &  --                        &  \(74\)

\\
\hline
\textit{total}                    &                            & \textbf{1,297}

\end{tabular}}

\caption{Frequency of validation errors in our
inter-annotator agreement study identified by \texttt{grewv}, broken down by type for errors of frequency $\geq$10. Annotations made after the introduction of our validation system would
forbid these disagreements. }
\label{tbl:iaa-grewv}
\vspace{-15pt}
\end{table}

\section{Technical information and reproducibility}\label{sec:tech}

The AlephBERT transformer model by \citet{seker-etal-2022-alephbert} used for most of the components in this paper is available from huggingface\footnote{\url{https://huggingface.co/models}} and was trained over approximately 8 days on a DGX machine (8 V100 GPUs) on close to 18 GB of text from the OSCAR corpus \cite{OrtizSuarezSagotRomary2019}, Hebrew Wikipedia and Hebrew Twitter. The model uses the standard 12 layer transformer with 768 dimensional word-piece representations or approximately 110 million parameters, with a vocabulary of 52K types. 

The flair taggers described above were trained with a biLSTM CRF (n\_hidden=256) on top of AlephBERT transformer word embeddings, concatenated with 5 dimensional MWT positional dense embeddings for tagging, and 17 dimensional POS embeddings as inputs for morphological features. In all cases we used the default flair hyperparameters with a learning rate of 0.1, optimizing with SGD using a mini-batch size of 15 and halting on dev set accuracy with a patience of 3.

The Diaparser model, also using default hyperparameters, combines fixed AlephBERT embeddings (no fine-tuning) with randomly initialized, fine-tunable fixed embeddings (100 dimensions), which are all fed into a 200-dimensional biLSTM topped by arc and dependency relation MLPs with biaffine attention (500 and 100 dimensional respectively), for a complete parser model with approximately 12M trainable parameters. All training was carried out on a consumer laptop (Dell XPS15, Intel Core i9-9980HK CPU@2.40GHz, 64 GB RAM, NVIDIA GeForce GTX 1650, 4GB GPU RAM).

For Stanza and Trankit's architectures, as well as RFTokenizer and Seker et al.'s system, we refer readers to the original published papers and tool documentations.

\section{Label distributions}\label{sec:labs} 

In compliance with the EMNLP reproducibility checklist, Table \ref{tab:labels} gives the exact breakdown of POS tag labels in each dataset.

\begin{table}[htb]
\centering
{\small %
\begin{tabular}{l|rr}
      & \textbf{\corpname{}}  & \textbf{HTB}   \\
      \hline
ADJ   & 1,711  & 1,256  \\
ADP   & 21,005 & 21,058 \\
ADV   & 1,529  & 1,035  \\
AUX   & 956   & 1,014  \\
CCONJ & 1,706  & 1,976  \\
DET   & 11,177 & 11,587 \\
INTJ  & 4     & 3     \\
NOUN  & 31,624 & 31,003 \\
NUM   & 1,126  & 1,220  \\
PRON  & 1,633  & 1,015  \\
PROPN & 11,448 & 1,181  \\
PUNCT & 11,613 & 11,301 \\
SCONJ & 1,317  & 1,792  \\
SYM   & 146   & 1     \\
VERB  & 11,650 & 11,359 \\
X     & 304   & 118  
\end{tabular}
}
\caption{POS label distributions.}\label{tab:labels}
\end{table}

\section{Confusion Matrices}
\label{sec:appendix}

The confusion matrices in Figure \ref{fig:confmat} below show predicted vs. gold label frequencies for POS tagging and dependency relations in the cross-corpus experiments using this paper's best system, HebPipe. All numbers are computed using the official CoNLL UD scorer and end-to-end predictions from plain text, taking the scorer's optimal alignment as the basis for matching predictions and gold token labels, which is non-trivial due to occasional differences in the predicted segmentation of words into tokens.

\begin{figure*}[htb]

\subfloat[train: Wiki $\rightarrow$ test: HTB]{%
  \includegraphics[clip,width=1\textwidth,trim = 0cm 0cm 0cm 1cm]{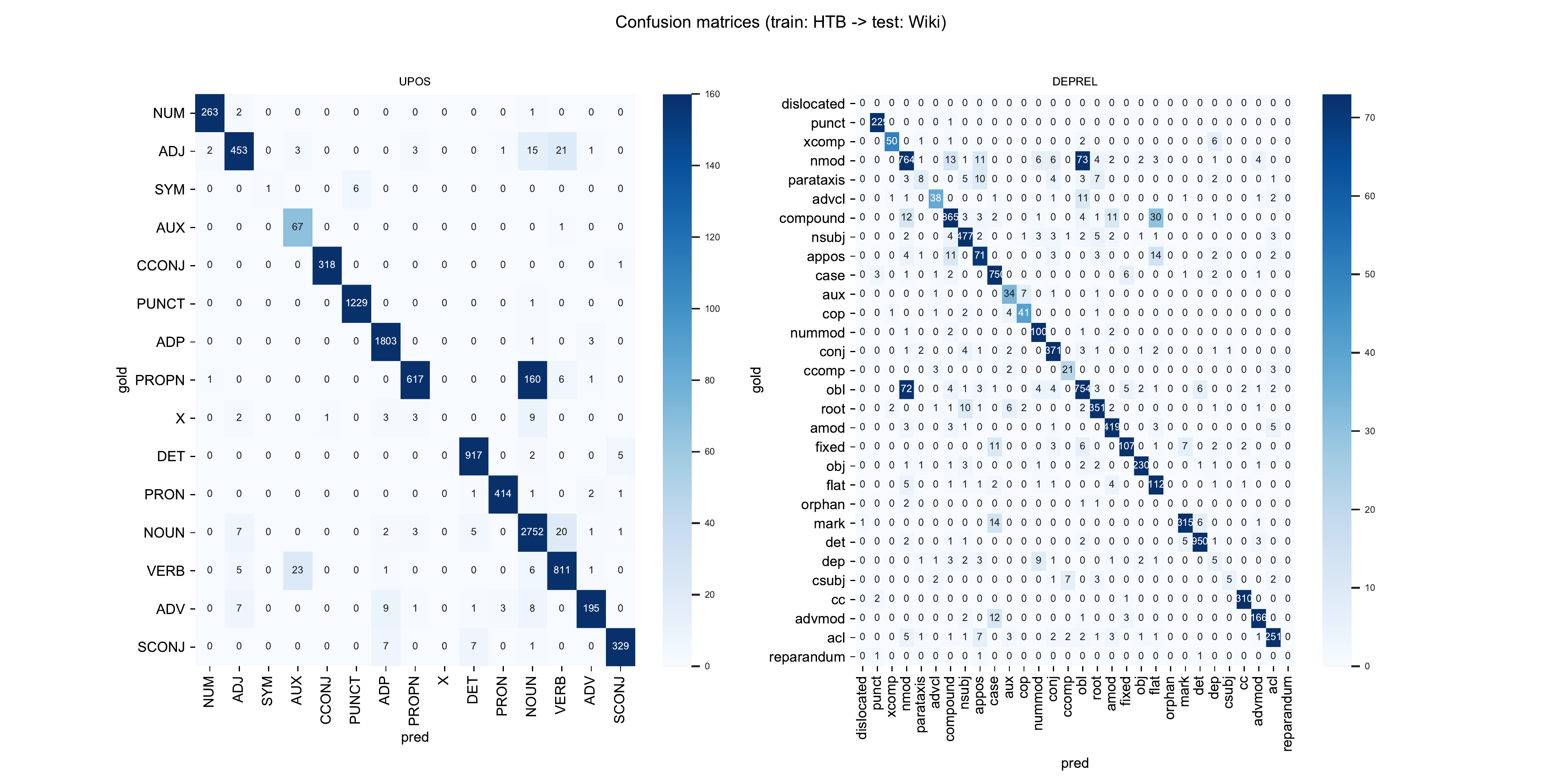}%
}

\subfloat[train: HTB $\rightarrow$ test: Wiki]{%
  \includegraphics[clip,width=1\textwidth,trim = 0cm 0cm 0cm 1.5cm]{confmat-htb-wiki.pdf}%
}

\caption{Confusion matrices for cross-corpus POS tag and dependency relation predictions in both directions. Top: training on HTB newswire and predicting on \corpname{}, bottom: training on \corpname{} and predicting on HTB.}\label{fig:confmat}

\end{figure*}

\end{document}